\documentclass[conference]{IEEEtran}
\IEEEoverridecommandlockouts
\usepackage{cite}
\usepackage{amsmath,amssymb,amsfonts}
\usepackage{graphicx}
\usepackage{textcomp}
\usepackage{xcolor}
\usepackage{url}
\usepackage{subcaption}
\usepackage{algorithm}
\usepackage{algorithmic} 

\usepackage{balance}

\def\BibTeX{{\rm B\kern-.05em{\sc i\kern-.025em b}\kern-.08em
    T\kern-.1667em\lower.7ex\hbox{E}\kern-.125emX}}

\usepackage{tikz}
\usepackage{textcomp}
\usepackage{hyperref}
\newcommand\copyrighttext{%
\footnotesize \textcopyright 2022 IEEE.
This paper has been accepted for presentation in
2nd International Conference on Computers and Automation (CompAuto 2022), August 18-20, 2022.
Personal use of this material is permitted. Permission from IEEE must be obtained for all other uses, in any current or future media, including reprinting/republishing this material for advertising or promotional purposes, creating new collective works, for resale or redistribution to servers or lists, or reuse of any copyrighted component of this work in other works. DOI: \href{<http://tex.stackexchange.com>}{<DOI No.>}}
\newcommand\copyrightnotice{%
\begin{tikzpicture}[remember picture,overlay]
\node[anchor=south,yshift=10pt] at (current page.south) {\fbox{\parbox{\dimexpr\textwidth-\fboxsep-\fboxrule\relax}{\copyrighttext}}};
\end{tikzpicture}%
}

\begin{document}

\title{Drones-aided Asset Maintenance in Hospitals

\thanks{This work was made possible by the Qatar National Research Fund (a member of Qatar Foundation) through the NPRP Grant under Grant NPRP12S-0313-190348. The statements made herein are solely the responsibility of the authors.}
}


\author{\IEEEauthorblockN{Muhammad Asif Khan}
\IEEEauthorblockA{\textit{Qatar Mobility Innovation Center (QMIC)} \\
\textit{Qatar University}\\
Doha, Qatar \\
asifk@ieee.org}
\and
\IEEEauthorblockN{Hamid Menouar}
\IEEEauthorblockA{\textit{Qatar Mobility Innovation Center (QMIC)} \\
\textit{Qatar University}\\
Doha, Qatar \\
hamidm@qmic.com}
\and
\IEEEauthorblockN{Ridha Hamila}
\IEEEauthorblockA{\textit{Electrical Engineering} \\
\textit{Qatar University}\\
Doha, Qatar \\
hamila@qu.edu.qa}
}

\maketitle
\copyrightnotice

\begin{abstract}
The rapid outbreak of COVID-19 pandemic invoked scientists and researchers to prepare the world for future disasters. During the pandemic, global authorities on healthcare urged the importance of disinfection of objects and surfaces. To implement efficient and safe disinfection services during the pandemic, robots have been utilized for indoor assets. In this paper, we envision the use of drones for disinfection of outdoor assets in hospitals and other facilities. Such heterogeneous assets may have different service demands (e.g., service time, quantity of the disinfectant material etc.), whereas drones have typically limited capacity (i.e., travel time, disinfectant carrying capacity). To serve all the facility assets in an efficient manner, the drone to assets allocation and drone travel routes must be optimized. In this paper, we formulate the capacitated vehicle routing problem (CVRP) to find optimal route for each drone such that the total service time is minimized, while simultaneously the drones meet the demands of each asset allocated to it. The problem is solved using mixed integer programming (MIP). As CVRP is an NP-hard problem, we propose a lightweight heuristic to achieve sub-optimal performance while reducing the time complexity in solving the problem involving a large number of assets.

\end{abstract}


\section{Introduction}
Maintenance planning and scheduling plays a key role in implementing an effective maintenance management program. An effective maintenance management program improves user experience by ensuring the timely completion of planned maintenance activities with minimum asset downtime and least interruption to facility users. It also aims at optimizing the resource utilization to reduce the overall maintenance cost. Over the last decade, the use of unmanned aerial vehicles (UAVs) aka drones has been introduced in many commercial applications. Drones can be used in performing various simpler maintenance activities that involves least human intervention and intelligence such as capturing photos to aid in conditional assessment of assets, automatic meter reading, cleaning of windows at high altitudes etc.

An interesting use case of drones in asset management is the disinfection of outdoor assets in a hospital. During the COVID-19 pandemic, the sensitization of assets has been highly encouraged by global authorities and health institutions such as Centers for Disease Control and Prevention (CDC) \cite{cdc} and World Health Organization (WHO) \cite{who}. As the COVID-19 viruses can easily transmit from people to surfaces and further to more people due to contact, the disinfection of assets such as elevators, benches, doors etc. is frequently required. While traditionally, disinfection is being done by trained cleaning professionals, there have been concerns that the cleaning staff and janitors are at high risks of being exposed to viruses during the pandemic \cite{cleaners}.

\begin{figure}[htbp]
    \centering
    \includegraphics[width=0.9\columnwidth]{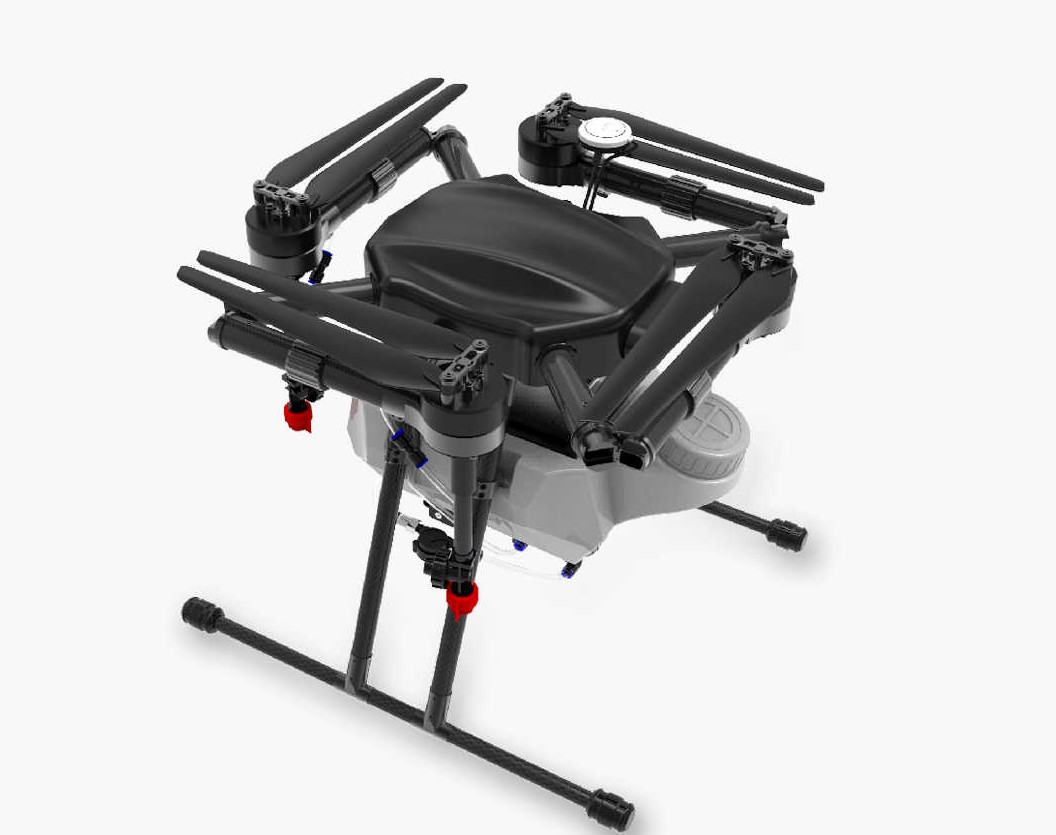}
    \caption{The MX450 spraying drone (image source - \cite{mx405}).}
    \label{fig:drone}
\end{figure}

To ensure the safety of people, facility users and cleaning professionals, service robots have been introduced to disinfect building indoors \cite{robotics, forbes}. However, when the assets are outdoors, robots might not be the capable to reach the heterogeneous types of assets. In this paper, we propose the use of drones for disinfection and sensitization of outdoor assets in hospitals such as benches, stretcher, wheel chairs, kids play areas, and ambulances etc. Drone-aided assets sensitization can be helpful to eliminate the risk of direct contact of the cleaning personnel with the COVID-19 viruses.

The recent advancements in drones technology have made it possible for use in applications such as crop-spraying in agriculture sector \cite{drones_spraying} and spraying for disinfection \cite{drones_spraying2}. Drone-aided disinfection services has has been introduced by US-based company EagleHawk \cite{eaglehawk}. Several drone models are designed for spraying applications such as Yamaha RMax \cite{RMax}, DJI Agras MG-1 \cite{agras}, MX405 etc. Fig. \ref{fig:drone} shows the MX405 spraying drone. The drone carries about 5.5 liters of materials (i.e., disinfectants) and has a maximum flight time of about $13$ minutes.

Given the limited capacity of spraying drones and short flight time operations, the drone operations must be optimized. In crop spraying, a drone can fly over a long stretch of crops area with continuous uninterrupted spraying. However, in applications such as spraying over particular assets located apart from each other, the travel time need to be highly optimized to reduce the unused flight time. \par

\begin{figure}[htp]
\centering
\includegraphics[width=0.95\columnwidth]{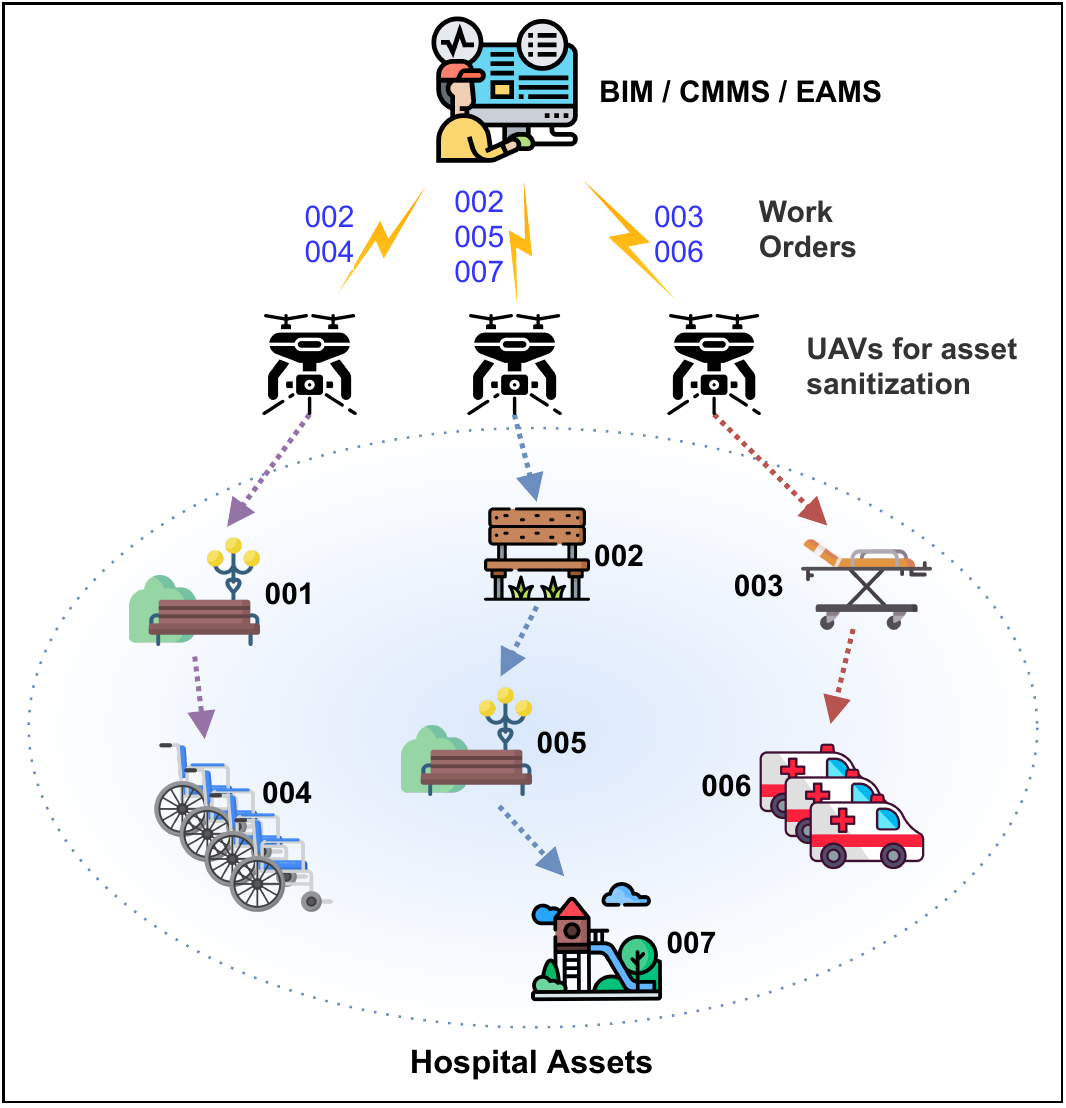}
\caption{Drone-based assets maintenance in hospitals.}
\label{fig:fig1}
\end{figure}

In a typical maintenance management setup, a planned preventive maintenance (PPM) plan is predefined in the computerized maintenance management system (CMMS) or sometimes refereed to as enterprise asset management system (EAMS) \cite{cmms}. A CMMS/EAMS is a local or cloud-based software system often integrated with building information management (BIM) system \cite{bim} that contains all the details of the asset including asset type, make and model, location coordinates, maintenance history, and upcoming maintenance schedule. The CMMS system auto-generates the maintenance work orders and send it to maintenance team for performing the activities. Fig. \ref{fig:fig1} illustrates the implementation of the aforementioned drone-aided maintenance management system.
\par
In this paper, we propose the use of drones for assets disinfection in critical facilities like hospitals. To improve the utilization of drones, we propose the optimal formulation of the problem for drone's path planning and assets allocation using capacitated vehicle routing problem (CVRP). Additionally, we propose a sub-optimal lightweight heuristic algorithm to solve the NP-hard CVRP problem in an efficient manner.

\section{System Model}
Consider a hospital area of 1000 $\times$ 1000 $m^2$ with a mix of assets. Let us consider there are four different types of assets i.e., benches, wheel chairs, ambulances, and kids playground equipment. The location coordinates of these assets are independently and uniformly distributed in the hospital area. We also have five drones to perform the maintenance of all these assets. We allocate these assets to each drone such that each asset is visited at most one time, and that each asset must be visited by at least one drone. The time a drone takes to reach an asset is is calculated from the distance between the current location of the drone to the asset given that the drone travels with constant speed of 20 meter per second \cite{drone_speeds}.

We define the cost of maintenance of an asset as the sum of travel time ($T_{ij} = \frac{d_{ij}}{V}$) to reach the asset (from drone's origin or current location) and the time required to sanitize the asset ($T_i$), where $d_{ij}$ represents the distance between drone's current location $i$ and asset location $j$, while $V$ is the constant speed of each drone.

\begin{equation} \label{eq:maint_cost}
    C_{ij} = T_{ij} + T_j
\end{equation}

Additionally, each asset also has a demand ($q_i$) i.e., the amount of sanitizer used for $i^{th}$ asset, where $q_0 = 0 for drones$. Typically, it is fixed for each asset type i.e., same amount of sanitizer is to be used for all benches. We also have a capacity constraint on each drone i.e., $C \geq q_i $. We assume that the demand for each asset is deterministic and unsplittable i.e., each drone must be exactly server by a single drone. We further assume that all drones are homogeneous i.e., all drones have same capacity. We also assume that the cost is symmetric i.e., $c_{ij} = c_{ji}$ .

\begin{figure*}[ht]
\centering
\subfloat[Distributed assets of multiple types.]{\includegraphics[width=8cm, height=8cm]{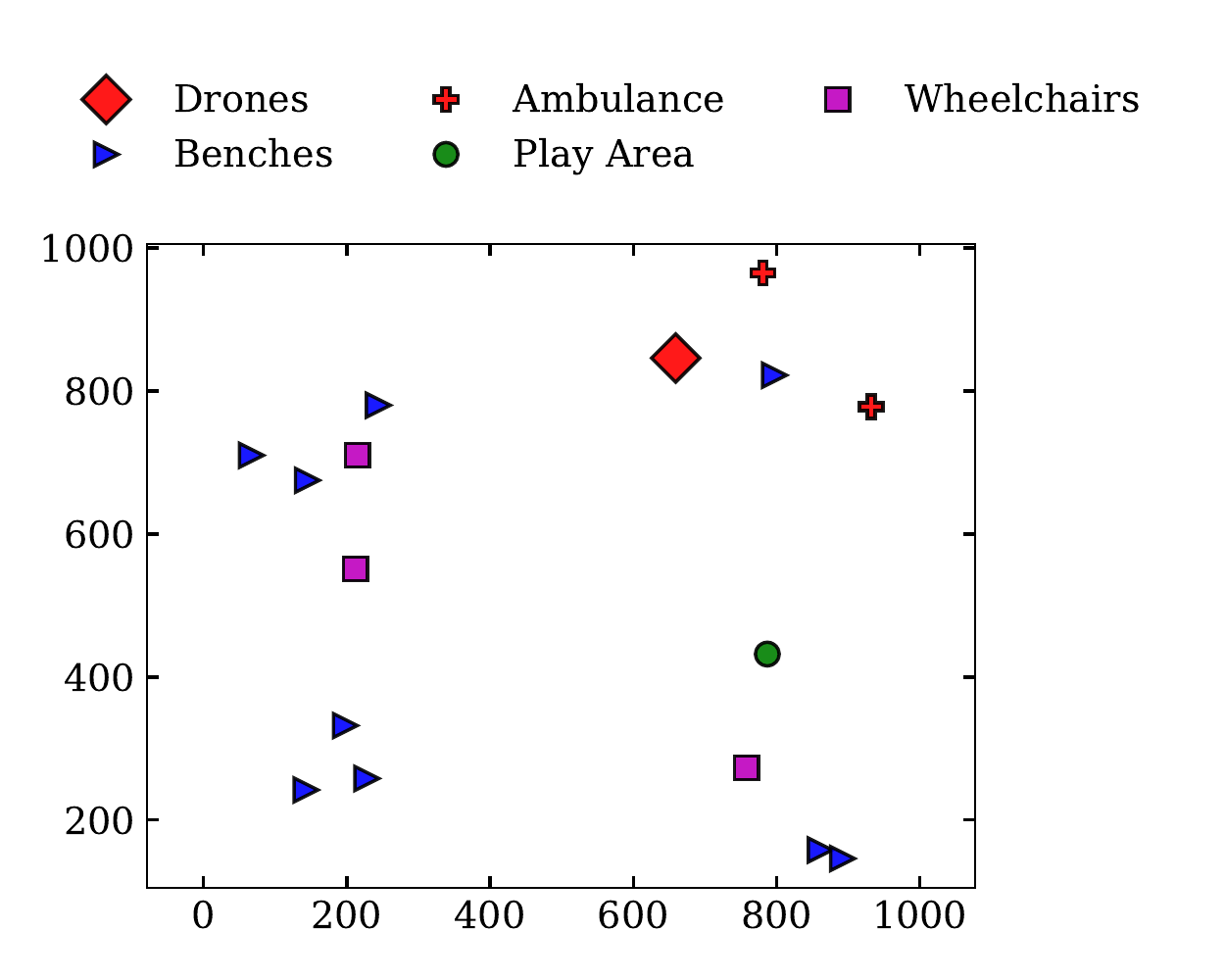}} \hspace{1em} \hspace{1.5em}
\subfloat[Optimal routes with 4 drones.]{\includegraphics[width=8cm, height=8cm]{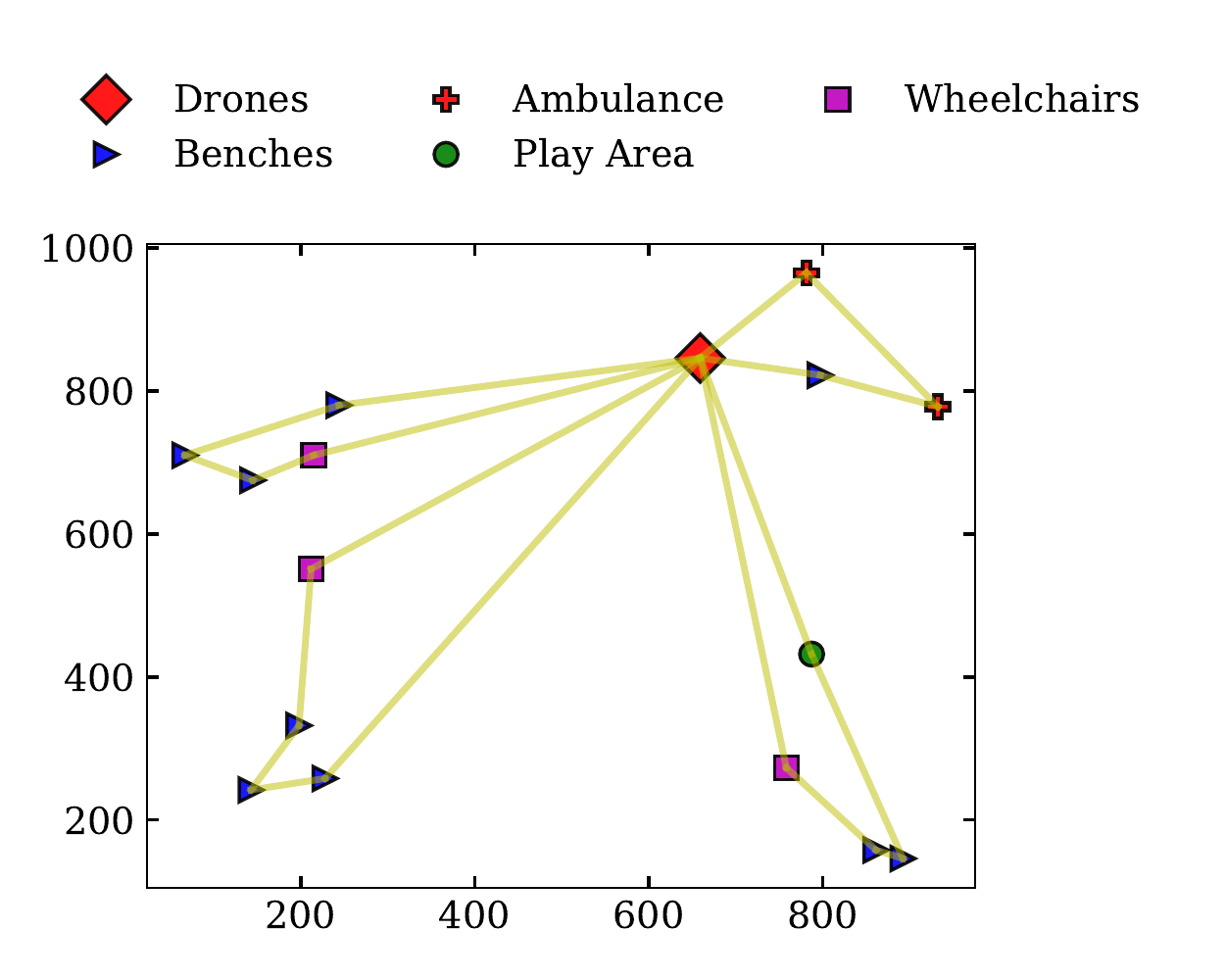}}
\caption{System model showing asset locations in 1000 $\times$ 1000 $m^2$ grid.}
\label{fig:fig2}
\end{figure*}

\subsection{Problem Formulation}
Let us say that $n$ is the total number of assets of all types, $m$ is the number of drones to serve the assets. We define $G(V,A)$ a complete graph such that $V={0,1,2,...n}$ is the set of nodes where, $0$ denotes the drone and ${1,2,...n}$ denotes the set of assets. We are interested to determine $m$ routes each travelled by a drone such that, (i) each route starts at $0$, (ii) each asset belongs to exactly one route, and (iii) the cumulative demand of all asset along each route must not exceed the capacity $C$.

We define a decision variable $x_{ij}$ as follow.
$$
x_{ij} =
\begin{cases}
1 & \text{if (i,j) belongs to the route.} \\
0 & \text{otherwise.}
\end{cases}
$$

\begin{equation} \label{eq:obj1}
\min \left \{ \sum\limits_{i \in V} \sum\limits_{j \in V} {C_{ij}x_{ij}} \right \} \tag{2}
\end{equation}

\begin{subequations}
\begin{align}
\text{subject to:} \notag \\
& \sum\limits_{i \in V}  x_{ij} = 1  \quad \forall j \in V \ \{0\} \label{eq:obj1_c1} \\
& \sum\limits_{j \in V}   x_{ij} = 1  \quad \forall i \in V \ \{0\} \label{eq:obj1_c2} \\
& \sum\limits_{i \in V} x_{i0} = m    \label{eq:obj1_c3} \\
& \sum\limits_{j \in V} x_{0j} = m    \label{eq:obj1_c4} \\
& x_{ij} \in {0,1}  \quad \forall i \in V, j \in V  \label{eq:obj1_c5} \\
& u_i - u_j + Q x_{ij} \leq  Q - q_i \quad \forall  i,j \in V \ \{0\} \label{eq:obj1_c6} \\
& q_i \leq u_i \leq Q \quad \forall i \in V \ \{0\} \label{eq:obj1_c7}
\end{align} 
\end{subequations}

The objective function (Eq. \ref{eq:obj1}) aims to minimize the total travel time of all drones serving the entire assets.
Constraints (\ref{eq:obj1_c1} and \ref{eq:obj1_c2}) ensure that each asset be exactly visited by one drone.
Constraints (\ref{eq:obj1_c3} and \ref{eq:obj1_c4}) ensure that exactly $m$ routes to be created.
Constraints (\ref{eq:obj1_c5} is integer constraint on the decision variable $x_{ij}^k$.
Constraints (\ref{eq:obj1_c6} and \ref{eq:obj1_c7}) are known as Miller, Tucker and Zemlin (MTZ) constraints \cite{MTZ_constraints}. The auxiliary variable $u_i$ represents the remaining capacity of the drone after serving asset $i$. Constraint \ref{eq:obj1_c6} particularly can be explained in this way. If $x_{ij}=0$, no drone travels from $i$ to $j$, the constraint is always true and thus this constraint becomes redundant. If $x_{ij}=1$, a drone travels from $i$ to $j$ along the path $(i,j)$ and its capacity after serving $j$ becomes $\geq u_i + q_j$. Constraint \ref{eq:obj1_c6} is the capacity constraint which means that the remaining capacity at any asset along the route can never exceed the drone capacity, whereas it must fulfill the demand of the asset $j$ being visited by the drone.

The problem formulation stated in Eq. \ref{eq:obj1}, with constraints \ref{eq:obj1_c1}-\ref{eq:obj1_c5} is known as capacitated vehicle routing problem (CVRP) \cite{cvrp1, cvrp2}. The two additional MTZ constraints in Eq. \ref{eq:obj1_c6}-\ref{eq:obj1_c7} are used to eliminate sub-tours and impose capacity of the drones.
Fig. \ref{fig:fig2} illustrates a simple scenario of randomly located assets with optimal routes from the drone's original location to reach assets using the optimal CVRP scheme.

\subsection{Heuristic: Greedy Nearest Neighbor (GNN)} \label{subsec:h1}
The proposed CVRP problem is NP-hard. There exists several heuristics solutions to such problems such as greedy search, simulated annealing, genetic algorithm, and deep reinforcement learning (DRL) \cite{cvrp_drl2021}. However, we are interested to find a lightweight heuristic that can provide sub-optimal performance. We propose a heuristic called as "Greedy Nearest Neighbor (GNN)" algorithm to accomplish the task. Our heuristic is based on the same assumptions used in the optimal CVRP scheme and does not take any extra assumption. 
The GNN heuristic is stated in Algorithm \ref{alg:gnn}. It is divided into two phases. In the first phase, the network is divided into $K$ clusters (where $K=m$ the number of available drones) such that the cumulative demand of each cluster does not exceed the drone capacity, while the total cost (i.e., travel time) is minimized. The clustering problem is formulated using integer linear programming (ILP) as follow:

We define two decision variables as $x_{ij}$ and $y_i$.
$$ x_{ij} = 1 \; \text{(if point $j$ is connected to median $i$.)} $$
$$ y_{i} =  1 \; \text{(if point $i$ is chosen as median.)} $$

\begin{equation} \label{eq:obj2}
\min \left \{ \sum\limits_{i \in V} \sum\limits_{j \in V} {C_{ij}x_{ij}} \right \} \tag{3}
\end{equation}

\begin{subequations}
\begin{align}
\text{subject to:} \notag \\
& \sum\limits_{j \in V} x_{ij} = 1  \quad \forall i \in V \ \{0\} \label{eq:obj2_c1} \\
& \sum\limits_{j \in V} y_{i} = m    \label{eq:obj2_c2} \\
& \sum\limits_{j \in V} x_{ij} = y_i    \label{eq:obj2_c3}
\end{align} 
\end{subequations}

The aforementioned ILP problem can be solved in polynomial time. In the next phase

\algsetup{indent=1em}
\setcounter{algorithm}{-1}
\begin{algorithm}[h]
\caption{Calculate $y = x^n$}
\label{algo:heuristic}
\begin{algorithmic}[1]
 \STATE{\textbf{Inputs:} ($n, m, q, Q, loc, S$)}
 \STATE{\textbf{Outputs:} (m routes)}
 
 \STATE {Find distance matrix from coordinates $loc$} \\
 $d_{ij} = euclidean(loc[:,0], loc[:,1])$
 \STATE {Find the travel time matrix from the distance matrix.} \\
 $T_{ij} = d_{ij} / S$
 \\[0.5em]
 
 \STATE {\textbf{Phase 1:} \textit{Clustering}}
 \STATE {Divide assets into K disjoint clusters.}
 \STATE {Find $K$ medians with $S_i^j$ i.e., set $j$ connected to median $i^{th}$.}
 \\[0.5em]
 
 \STATE {\textbf{Phase 2:} \textit{Routing}}
 \FOR{$i \in K$}
    \FOR {$j \in S$}
     \STATE {Sort $d_{ij}$ $\forall i \in S j \in S$ into descending order.}
     \STATE {Route drone to median: $0 \rightarrow i$}
     \STATE {Route from median to all locations in the cluster: $i \rightarrow j$ to}
    \ENDFOR
 \ENDFOR

\caption{Greedy Nearest Neighbor (GNN).}
\label{alg:gnn}
\end{algorithmic}
\end{algorithm}

In the next phase, each cluster is assigned to one drone. As the drones have equal capacities, any drone-cluster assignment can be used as the assignment does not impact the system cost. Then for each cluster, the travel costs for all edges $(i,j)$ are sorted in ascending order. The assigned drone is moved to the nearest node in the cluster and then route through each edge in the list. When it reaches last node, it return to the origin position.

\section{Performance Evaluation}
\begin{figure}[!h]
    \centering
    \includegraphics[width=0.95\columnwidth]{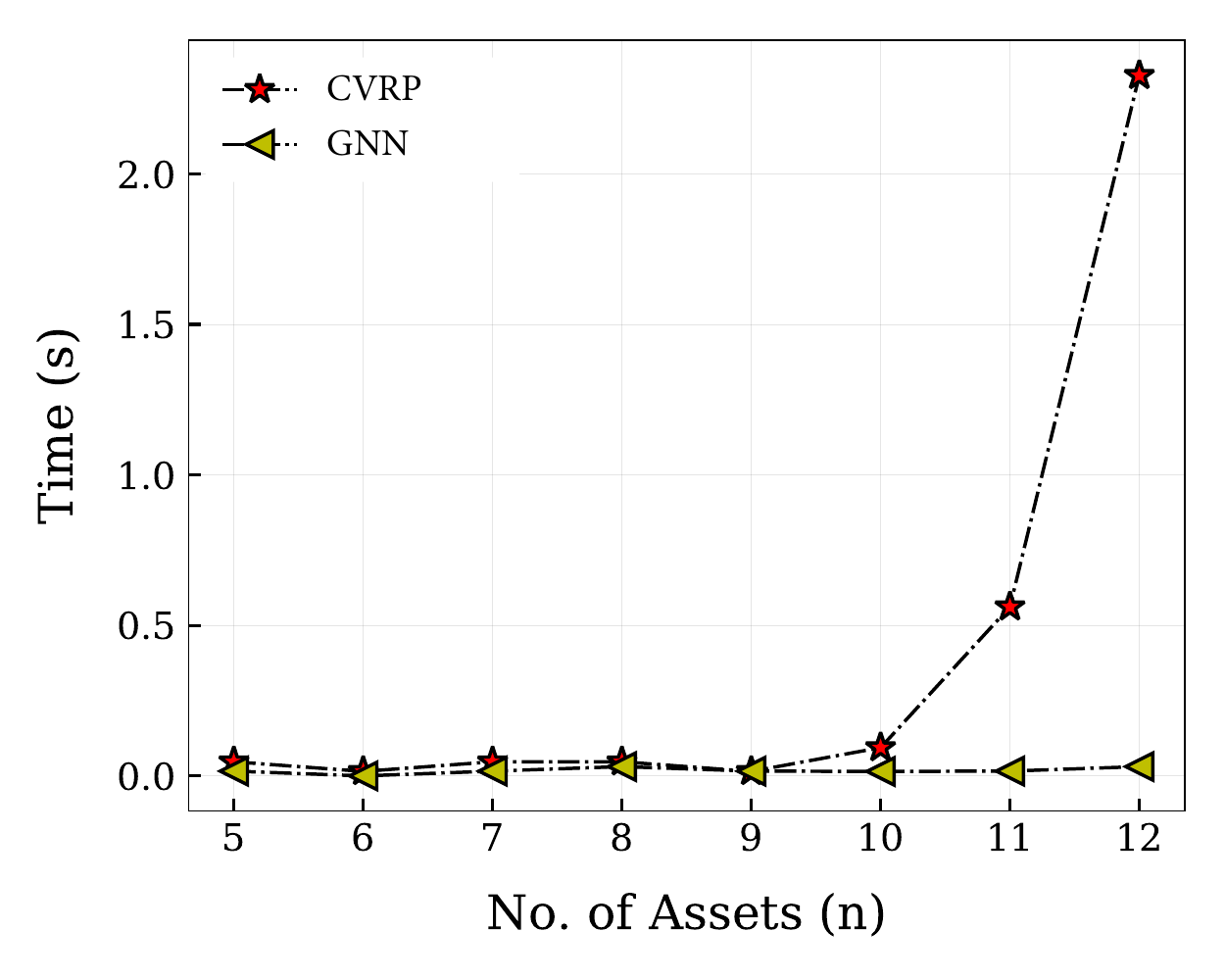}
    \caption{Time complexity comparison: Optimal scheme versus GNN heuristic.}
    \label{fig:fig3}
\end{figure}

The proposed optimal scheme and GNN heuristic are implemented using CVXPY \cite{cvxpy} to evaluate the performance. CVXPY is a modeling language for convex optimization problems developed at Stanford university. First, a comparison of the time complexity of the optimal CVRP and sub-optimal GNN heuristic is presented in Fig. \ref{fig:fig3}. It is worthy to note that the purpose of the GNN heuristic was to reduce the time complexity involved with the large problems when the network size increases.

It can be observed that the computational time of the GNN heuristic is significantly lower than the optimal CVRP scheme. The CVRP is a known NP-hard problem and the complexity increases exponentially with the increasing size of the network. Comparatively, the GNN heuristic solve the problem into two phases (i.e., cluster and route) that leads to lower computational time.

\begin{figure}[!h]
    \centering
    \includegraphics[width=0.95\columnwidth]{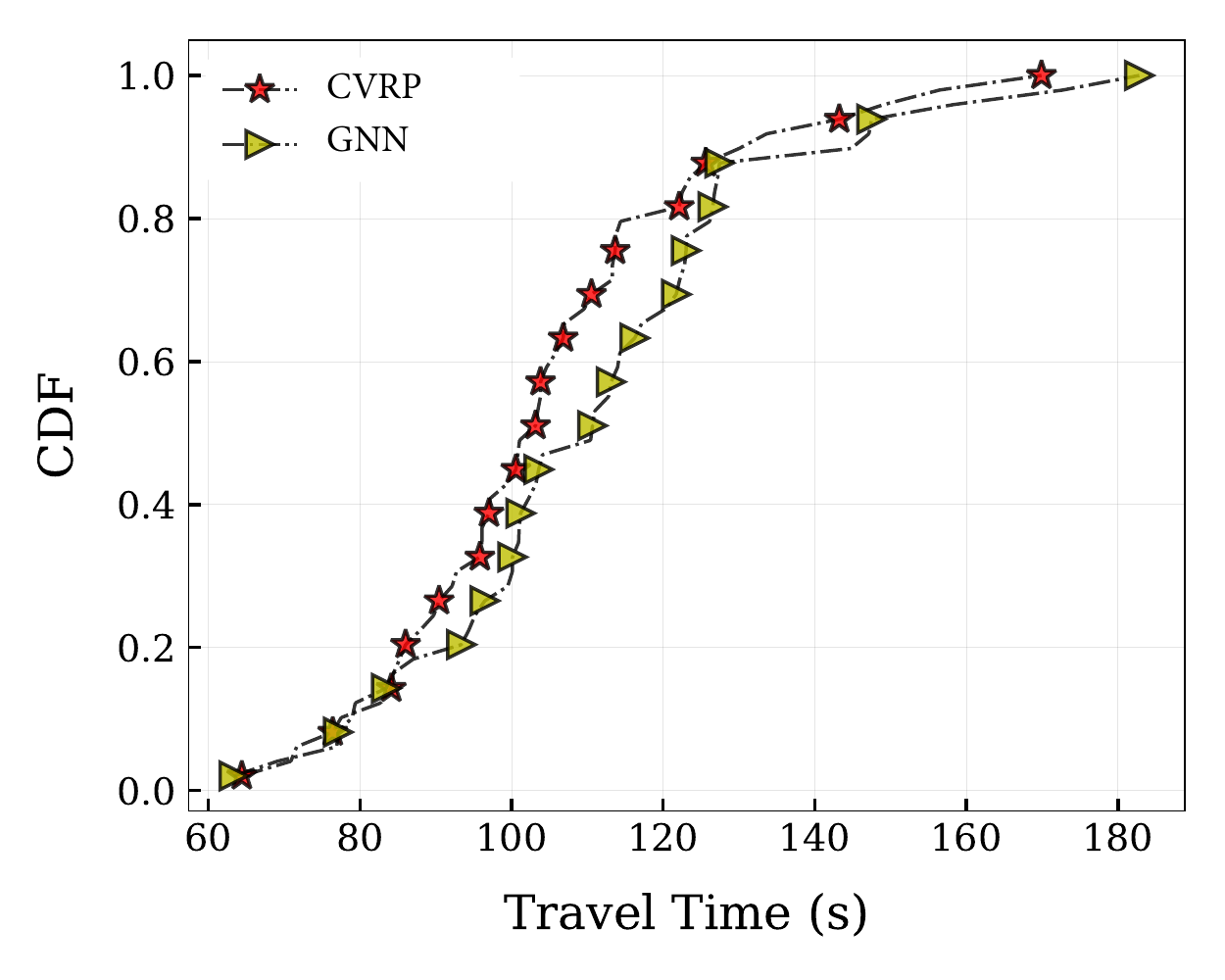}
    \caption{Travel time comparison for optimal scheme versus GNN heuristic.}
    \label{fig:fig4}
\end{figure}

Next, a comparison of both schemes to achieve the performance goal (i.e., travel time minimization) is performed using fixed network size over several simulation runs (10-100 runs for different experiments). In each run, the network of assets is created with random locations of assets and the optimal routes are found using both schemes. The travel time comparison is depicted in Fig. \ref{fig:fig4} using cumulative distribution function (CDF) plot. 

It can be observed that the GNN heuristic always achieves close to optimal performance. \par

\begin{figure}[!h]
    \centering
    \includegraphics[width=0.95\columnwidth]{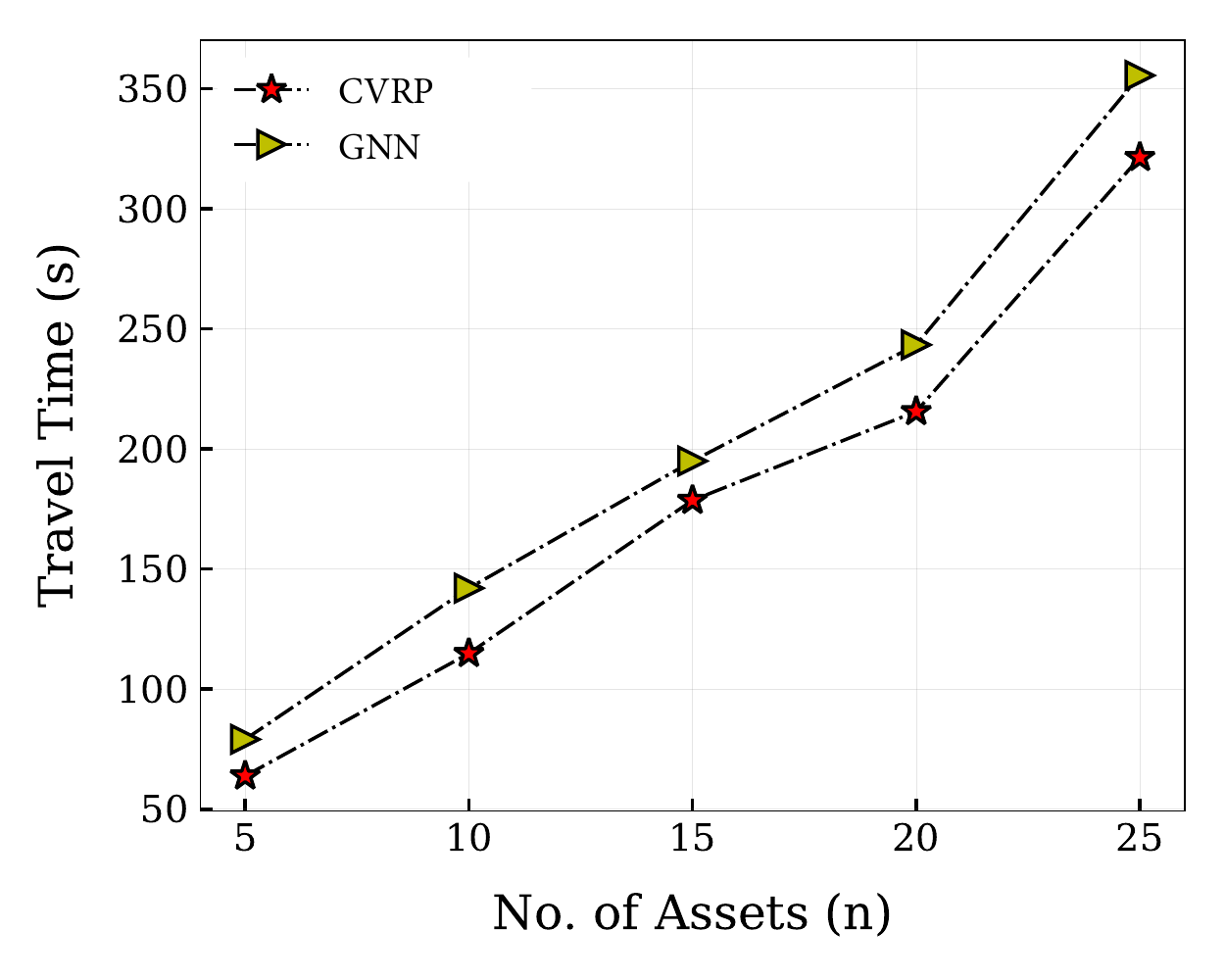}
    \caption{Travel time comparison: Optimal scheme versus GNN heuristic.}
    \label{fig:fig5}
\end{figure}

We are also interested to evaluate the two schemes for performance gains over various network scales. We further compared the two schemes by varying the number of assets in the network from $10-30$ to solve the routing problem and computing the travel time. Fig. \ref{fig:fig5} depicts the performance comparison of both schemes. The results further validates the performance gains in Fig. \ref{fig:fig4} and the GNN scheme achieves close to optimal performance by finding the sub-optimal routes.

\section{Conclusion}
This paper proposes a drone-aided asset disinfection service for large facilities such as hospitals. Two algorithms i.e., CVRP (optimal but NP-hard) and a GNN heuristic (sub-optimal but lightweight) are proposed to solve the problem of finding shortest disjoint routes for drones covering all assets while meeting the demands of all assets and efficiently utilizing the capacity of the drones. The performance evaluation shows that the GNN heuristic provides close to optimal performance and can be efficiently used to solve large-size CVRP problems. The proposed scheme can be equally adopted for other similar applications such as asset condition surveys and drone-aided automatic meter readings. Future extension of this work include scenarios that take into account asset unavailability (e.g., occupancy), and deterministic service time window requirements.


\balance
\bibliographystyle{ieeetr} 
\bibliography{biblio}

\begin{thebibliography}{10}

\bibitem{cdc}
{Disease Control and Prevention (CDC)}, ``{Cleaning and Disinfecting Your
  Facility}.''
  \url{https://www.cdc.gov/coronavirus/2019-ncov/community/disinfecting-building-facility.html}.
\newblock Online; accessed 16 January 2022.

\bibitem{who}
{World Health Organization (WHO)}, ``{Getting your workplace ready for
  COVID-19}.''
  \url{https://www.who.int/docs/default-source/coronaviruse/getting-workplace-ready-for-covid-19.pdf/}.
\newblock Online; accessed 23 January 2022.

\bibitem{cleaners}
{OHS Online}, ``{Cleaning Workers are on the Front Lines of the Coronavirus
  Pandemic}.''
  \url{https://ohsonline.com/Articles/2020/03/20/Cleaning-Workers-are-on-the-Front-Lines-of-the-Coronavirus-Pandemic.aspx?Page=1&m=1}.
\newblock Online; accessed 23 January 2022.

\bibitem{mx405}
``{MX405 Portable Ulv Electric Sprayer Sprayer Atomizer Garden Motor Operated
  Pesticide Agricultural Sprayer Drones (MX405 White)}.''
  \url{https://en.asuav.com/show-2059.html}.
\newblock Online; accessed 23 January 2022.

\bibitem{robotics}
J.~Holland, L.~Kingston, C.~McCarthy, E.~Armstrong, P.~O’Dwyer, F.~Merz, and
  M.~McConnell, ``Service robots in the healthcare sector,'' {\em Robotics},
  vol.~10, no.~1, 2021.

\bibitem{forbes}
{Rich Blake}, ``{In Coronavirus Fight, Robots Report For Disinfection Duty}.''
  \url{https://www.forbes.com/sites/richblake1/2020/04/17/in-covid-19-fight-robots-report-for-disinfection-duty/?sh=6d7fcc1b2ada/}.
\newblock Online; accessed 23 January 2022.

\bibitem{drones_spraying}
M.~Hentschke, E.~Pignaton~de Freitas, C.~H. Hennig, and I.~C. Girardi~da Veiga,
  ``Evaluation of altitude sensors for a crop spraying drone,'' {\em Drones},
  vol.~2, no.~3, 2018.

\bibitem{drones_spraying2}
H.~González~Jorge, L.~M. González~de Santos, N.~Fariñas~Álvarez,
  J.~Martínez~Sánchez, and F.~Navarro~Medina, ``Operational study of drone
  spraying application for the disinfection of surfaces against the covid-19
  pandemic,'' {\em Drones}, vol.~5, no.~1, 2021.

\bibitem{eaglehawk}
{InterDrone}, ``{EagleHawk provides effective solution for disinfecting large
  venues against COVID-19}.''
  \url{https://interdrone.com/interdrone-interview/eaglehawk-provides-effective-solution-for-disinfecting-large-venues-against-covid-19/}.
\newblock Online; accessed 23 January 2022.

\bibitem{RMax}
``{Yamaha RMAX Helicopter}.''
  \url{https://www.yamahamotorsports.com/motorsports/pages/precision-agriculture-rmax}.
\newblock Online; accessed 23 January 2022.

\bibitem{agras}
``{DJI Agras MG-1}.'' \url{https://www.dji.com/mg-1}.
\newblock Online; accessed 23 January 2022.

\bibitem{cmms}
M.~Wienker, K.~Henderson, and J.~Volkerts, ``The computerized maintenance
  management system an essential tool for world class maintenance,'' {\em
  Procedia Engineering}, vol.~138, pp.~413--420, 2016.
\newblock SYMPHOS 2015 - 3rd International Symposium on Innovation and
  Technology in the Phosphate Industry.

\bibitem{bim}
R.~Volk, J.~Stengel, and F.~Schultmann, ``Building information modeling (bim)
  for existing buildings — literature review and future needs,'' {\em
  Automation in Construction}, vol.~38, pp.~109--127, 2014.

\bibitem{drone_speeds}
{Nick Cast}, ``{How Fast Can a Drone Fly? Updated 2021}.''
  \url{https://www.remoteflyer.com/how-fast-can-a-drone-fly-with-examples/}.
\newblock Online; accessed 16 January 2022.

\bibitem{MTZ_constraints}
T.~Sawik, ``A note on the miller-tucker-zemlin model for the asymmetric
  traveling salesman problem,'' {\em Bulletin of the Polish Academy of
  Sciences: Technical Sciences}, vol.~64, no.~No 3, pp.~517--520, 2016.

\bibitem{cvrp1}
Z.~Borcinova, ``Two models of the capacitated vehicle routing problem,'' {\em
  Croatian Operational Research Review}, vol.~8, pp.~463--469, 12 2017.

\bibitem{cvrp2}
T.~Ralphs, L.~Kopman, W.~Pulleyblank, and L.~Trotter, ``On the capacitated
  vehicle routing problem,'' {\em Mathematical Programming}, vol.~94,
  pp.~343--359, 01 2003.

\bibitem{cvrp_drl2021}
J.~Li, Y.~Ma, R.~Gao, Z.~Cao, A.~Lim, W.~Song, and J.~Zhang, ``Deep
  reinforcement learning for solving the heterogeneous capacitated vehicle
  routing problem,'' {\em IEEE Transactions on Cybernetics}, p.~1–14, 2021.

\bibitem{cvxpy}
S.~Diamond and S.~Boyd, ``{CVXPY}: A {P}ython-embedded modeling language for
  convex optimization,'' {\em Journal of Machine Learning Research}, vol.~17,
  no.~83, pp.~1--5, 2016.

\end{thebibliography}
\vspace{12pt}
\end{document}